\documentclass[10pt,twocolumn,letterpaper]{article}

\usepackage{cvpr}
\usepackage{times}
\usepackage{epsfig}
\usepackage{graphicx}
\usepackage{amsmath}
\usepackage{amssymb}
\usepackage{subfigure}
\usepackage{algorithmic}
\usepackage{algorithm}

\usepackage{multirow}
\usepackage{tabularx}
\usepackage{booktabs}
\graphicspath{{./figure/}}

\usepackage{authblk}
\usepackage[pagebackref=true,breaklinks=true,letterpaper=true,colorlinks,bookmarks=false]{hyperref}

\cvprfinalcopy 

\def\cvprPaperID{} %

\makeatletter
\newcommand{\printfnsymbol}[1]{%
  \textsuperscript{\@fnsymbol{#1}}%
}

\ifcvprfinal\fi
\begin{document}

\title{K-FACE: A Large-Scale KIST Face Database in Consideration with Unconstrained Environments}

\author[1,2]{Yeji Choi\thanks{Equal contribution (cyjcyj91@yonsei.ac.kr, phj0307@kist.re.kr)}}
\author[1,3]{Hyunjung Park \printfnsymbol{1}}
\author[1]{Gi Pyo Nam}
\author[1]{Haksub Kim}
\author[1]{Heeseung Choi}
\author[1]{Junghyun Cho}
\author[1,3]{Ig-Jae Kim\thanks{Corresponding author (drjay@kist.re.kr)}}
\affil[1]{Korea Institute of Science and Technology (KIST)}
\affil[2]{Yonsei University}
\affil[3]{KIST-School, University of Science and Technology}

\renewcommand\Authands{ and }

\maketitle

\begin{abstract}
In this paper, we introduce a new large-scale face database from KIST, denoted as K-FACE, and describe a novel capturing device specifically designed to obtain the data. The K-FACE database contains more than 1 million high-quality images of 1,000 subjects selected by considering the ratio of gender and age groups. It includes a variety of attributes, including 27 poses, 35 lighting conditions, three expressions, and occlusions by the combination of five types of accessories. As the K-FACE database is systematically constructed through a hemispherical capturing system with elaborate lighting control and multiple cameras, it is possible to accurately analyze the effects of factors that cause performance degradation, such as poses, lighting changes, and accessories. We consider not only the balance of external environmental factors, such as pose and lighting, but also the balance of personal characteristics, such as gender and age group. The gender ratio is the same, while the age groups of subjects are uniformly distributed from the 20s to 50s for both genders. The K-FACE database can be extensively utilized in various vision tasks, such as face recognition, face frontalization, illumination normalization, face age estimation, and three-dimensional face model generation. We expect systematic diversity and uniformity of the K-FACE database to promote these research fields.

\end{abstract}


\begin{table*}

\renewcommand{\arraystretch}{1.2}
\centering
  \caption{Comparisons of public databases.}
  \label{table:summary_of_database}
  
{\footnotesize
\begin{tabularx}{\linewidth}{X*{10}{c}}\toprule
\multirow{2}{*}{\textbf{Database}} & \multirow{2}{*}{\textbf{\# Subjects}} & \multirow{2}{*}{\textbf{\# Images}} & \multirow{2}{*}{\textbf{Pose}} & \multicolumn{2}{c}{\textbf{Illumination}} & \multirow{2}{*}{\textbf{Expression}} & \multirow{2}{*}{\textbf{Accessory}} & \multirow{2}{*}{\textbf{Age}} & \multirow{2}{*}{\textbf{Gender}}
\\ \cline{5-6} 
 & & & & \textbf{Type} & \textbf{Intensity}
\\ \midrule
\multirow{2}{*}{CelebA~\cite{CelebA}} & \multirow{2}{*}{10,177}    & \multirow{2}{*}{202,599}   & \multirow{2}{*}{No label}  & \multirow{2}{*}{No label}  & \multirow{2}{*}{No label}  & \multirow{2}{*}{No label}  & \multirow{2}{*}{No label}  & \multirow{2}{*}{No label}  & Labeled \\ 
&&&&&&&&&(58.0\% females) \\

CASIA- & \multirow{2}{*}{10,575}    & \multirow{2}{*}{494,414}   & \multirow{2}{*}{No label}  & \multirow{2}{*}{No label}  & \multirow{2}{*}{No label}  & \multirow{2}{*}{No label}  & \multirow{2}{*}{No label}  & \multirow{2}{*}{No label}  & Labeled \\
WebFace~\cite{CASIA_Web}&&&&&&&&&(41.1\% females)
\\
\multirow{2}{*}{VGGFace2~\cite{VGGFace2}} & \multirow{2}{*}{9,131}    & \multirow{2}{*}{3,310,000}   & \multirow{2}{*}{3}  & \multirow{2}{*}{No label}  & \multirow{2}{*}{No label}  & \multirow{2}{*}{No label}  & \multirow{2}{*}{No label}  & Labeled   & Labeled \\
&&&&&&&&(Young or mature)  &(40.7\% females)
\\
\multirow{2}{*}{Multi-PIE~\cite{MultiPIE}} & \multirow{2}{*}{337}   & \multirow{2}{*}{755,307}   & \multirow{2}{*}{15}  & \multirow{2}{*}{19}  & \multirow{2}{*}{No label}  & \multirow{2}{*}{6}  & \multirow{2}{*}{$\times$}  & Labeled  & Labeled \\
&&&&&&&&(Not balanced) & (30.3\% females)
\\
\multirow{2}{*}{M$^2$FPA~\cite{M2FPA}} & \multirow{2}{*}{229}    & \multirow{2}{*}{397,544}   & \multirow{2}{*}{62}  & \multirow{2}{*}{7}  & \multirow{2}{*}{No label}  & \multirow{2}{*}{3}  & \multirow{2}{*}{1}  & \multirow{2}{*}{No label}  & \multirow{2}{*}{No label}
\\ 
\\
\midrule

\multirow{2}{*}{\textbf{K-FACE(Ours)}} & \multirow{2}{*}{1,000}   & \multirow{2}{*}{17,550,000}   & \multirow{2}{*}{27}  & \multirow{2}{*}{35}  & \multirow{2}{*}{0-1000 lux}  & \multirow{2}{*}{3}  & \multirow{2}{*}{6+}  & Labeled,  & Labeled\\
&&&&&&&&Balanced(20-59) &(50.2\% females)
\\ \toprule

\end{tabularx}
}
\end{table*}

\section{Introduction}

With the development of deep learning and the advent of large-scale facial databases, deep-learning-based algorithms have exhibited remarkable performances in challenging research topics of computer vision related to facial images, such as unconstrained face recognition, age estimation, and three-dimensional (3D) face transformation \cite{FR1,Age1,3D}. In particular, a large-scale face database is one of the most important factors for the success of deep learning models. The distinguishing features of a good database are high-quality images, number of subjects and images, as well as diversity of intra- and inter-class variations, uniformity of class distribution, and accurate annotation. However, most public face databases consider only some features and are not designed to focus on other variants. We address this issue by introducing a process of construction of a large-scale face database that guarantees the diversity and uniformity of data and accurate annotation. The proposed database enables to precise analysis of the effects of factors that cause performance degradation, such as poses, lighting changes, and accessories. 
\newline \indent Public large-scale unconstrained face databases are divided into two categories according to the method used for image acquisition, using web crawling and a capturing device. As shown in Table~\ref{table:summary_of_database}, the first approach to collect facial images of celebrities or public figures via the Internet, including CelebA~\cite{CelebA}, CASIA-WebFace~\cite{CASIA_Web}, and VGGFace2~\cite{VGGFace2}, can acquire a large number of identities. However, owing to the lack of capturing condition information, the process is time-consuming and complex to obtain a correct label for an image, and the obtained labels are limited to some categories and can be very unbalanced. For example, in VGG2, poses are categorized into only three angles (front, three-quarter, or profile views), while ages are classified into only two groups (deemed young or mature). The other approach, including Multi-PIE~\cite{MultiPIE} and M$^2$FPA~\cite{M2FPA}, collects face images using a specially manufactured capturing device by recruiting subjects. In this case, the number of identities is limited, but clear annotations can be arranged automatically. However, most of these databases focus solely on the environmental conditions during image acquisition, resulting in severely unbalanced gender and age distributions that are positively skewed toward younger subjects. Moreover, they do not sufficiently cover extreme pose variations, diverse lighting intensities and directions, or the presence of multiple accessories—all of which frequently occur in real-world scenarios. To address these limitations, we introduce a new large-scale, unconstrained face database, referred to as K-FACE. The main contributions of this study are summarized as follows.

\begin{itemize}
      \item \textbf{Large-scale and high quality.} 
      We construct the first large-scale and high-quality database of Koreans.
      We constructed the first large-scale, high-quality database of Koreans. It contains a total of 17,550,000 images of 1,000 subjects with 17,550 images per person with various attributes. All images were captured using digital single-lens reflex (DSLR) cameras, which provided high-quality images with resolutions of 2592$\times$1728. The proposed database has advantages in both the quantity of subjects and the quality of images. 
      \item \textbf{Diversity.} 
      Through the hemispherical capturing system consisting of 27 cameras and 10 lighting devices, we obtained face images including 27 poses with changes in roll and yaw directions, 35 lighting conditions according to combinations of intensity and direction, three expressions, and multiple occlusions with five types of accessories such as glasses, a cap, and a mask. This allows many researchers to advance their specific topic by sampling the most appropriate subset.
      \item \textbf{Well balanced.} 
      The proposed database includes all the above elements evenly according to gender and age. The construction of different types of subsets by controlling specific properties enables to configuration of various experimental environments and the judgment of the factors of degradation. 
\end{itemize}

The remainder of this paper is organized as follows.
Section II introduces several representative public face databases. Section III describes the details of the K-FACE settings and statistics. In Section IV, we present challenging research topics that can be addressed by utilizing the proposed database. Finally, Section V concludes the paper. \newline

\section{Related Studies}

\subsection{CelebA}

The CelebA database was introduced for face representation and attribute prediction in the wild~\cite{CelebA}~\cite{CelebA2}. All face images were collected via the Internet. It contains a total of 202,599 images from 10,177 identities with five landmark points and 40 face attributes. For the details of face attributes, it provides a binary attribute about specific factors, such as blonde hair and wearing sunglasses or not. In this regard, this database has been widely used not only for face recognition, but also for face landmark detection and attribute prediction. The research fields based on the CelebA database are extended to generative adversarial network-based methods, such as style transfer.

\subsection{CASIA-WebFace}

Yi \textit{et al.} introduced a novel database, CASIAWebFace database, which consists of 494,414 images from 10,575 individuals collected from the web~\cite{CASIA_Web}. There are no environmental constraints in the collected image, such as pose, illumination, and expression. This implies that the CASIAWebFace database is close to the real-world environment, so that the database has been widely used to train unconstrained face recognition models. However, it is challenging to analyze the factors that affect the recognition performance because they do not provide the attribute information of the data, such as pose direction and illumination condition.

\subsection{VGGFace2}

The VGGFace database, the first version of VGGFace2, was released by Parkhi \textit{et al.}. in 2015~\cite{VGGFace}. This database aims to train for the increase in face recognition accuracy. In terms of face recognition, the number of identities and images of each subject in the training set is one of the most important factors affecting the recognition performance. To consider this issue, the VGGFace database consists of 2.6 million images acquired from 2,622 people by crawling the web. Cao \textit{et al.} introduced the VGGFace2~\cite{VGGFace2} database, which was extended from VGGFace. It is constructed from 9,131 people and has a total of 3.31 million images. For ethnic balancing, it includes more Asian people than the previous version, although it is still limited. In addition, the database provides annotations about pose and age.

\subsection{Multi-PIE}

Gross \textit{et al.} released a face database, denoted as Multi-PIE, acquired under various camera directions and illumination conditions to consider the pose, illumination, and expression variation~\cite{MultiPIE}. It includes over 750,000 images captured from 337 individuals and considers a total of 15 camera directions and 18 light directions. Thirteen cameras were located at head height with 15$^{\circ}$ intervals, from -90$^{\circ}$ to +90$^{\circ}$, and the others were installed above the subjects in consideration of the surveillance camera view. Illuminators were also installed near the locations of the cameras to consider various directions. However, it does not consider the luminance power; therefore, all images have the same lighting power.

\subsection{M$^2$FPA}

Li \textit{et al.} released the M$^2$FPA database, which addresses the facial pose analysis, including frontalization, pose estimation, and pose-invariant face recognition~\cite{M2FPA}. It contains 397,544 images of 229 subjects with yaw, pitch, attribute, illumination, and accessory. The authors designed an image acquisition system to capture the person in multiple yaw and pitch directions. They evaluated the performance of face frontalization and pose-invariant face recognition on M2FPA based on several state-of-the-art methods, such as DR-GAN~\cite{DRGAN} and TP-GAN~\cite{TPGAN}.

\section{K-FACE Database}

\begin{figure}[h]
  \centering
  \includegraphics[width=8cm]{{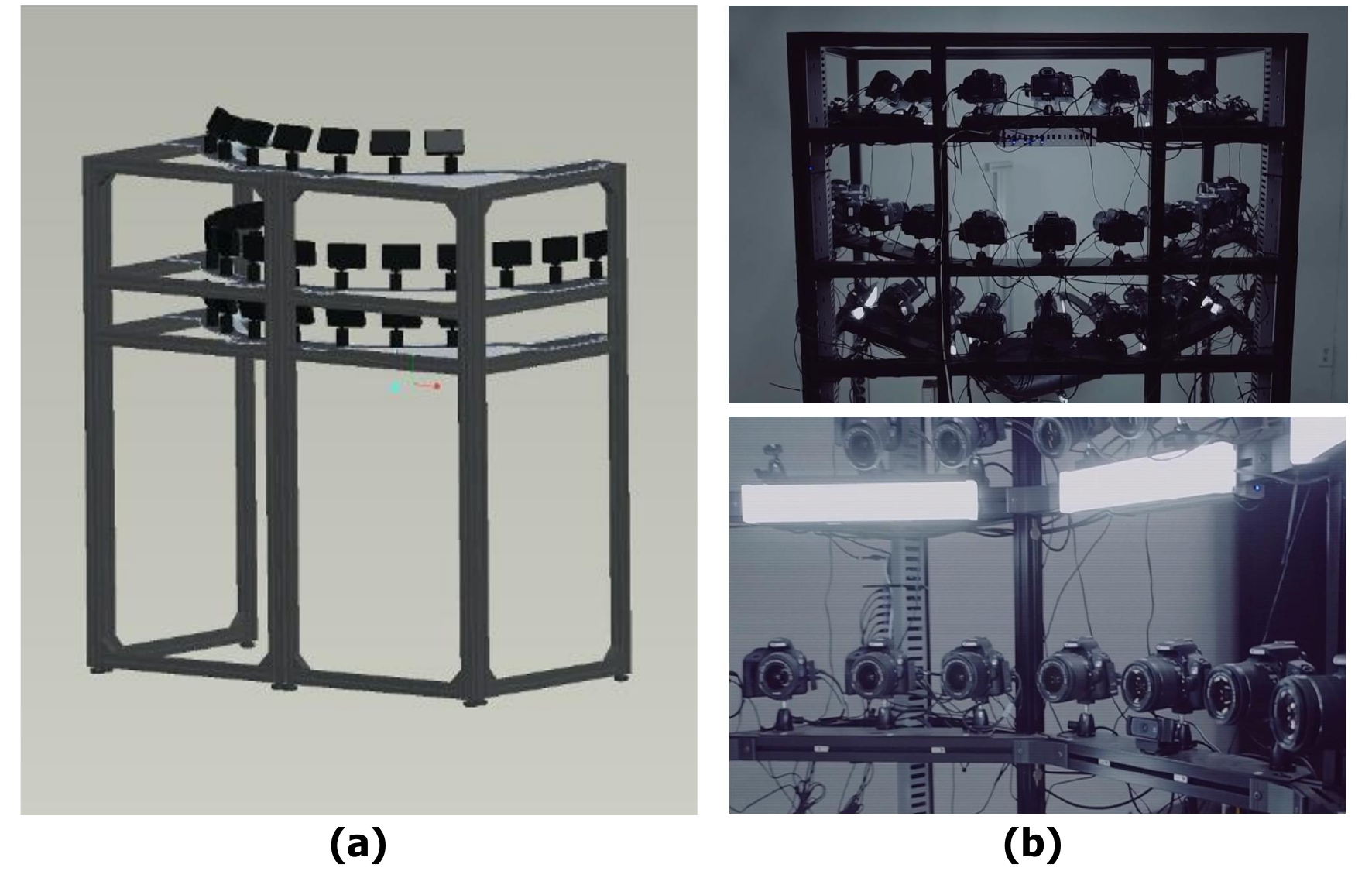}}
  \caption{Capturing device. (a) Concept design and (b) actual appearance. }
  \label{fig:device_concept}
\end{figure}

\begin{figure}[h]
  \centering
  \includegraphics[width=9cm]{{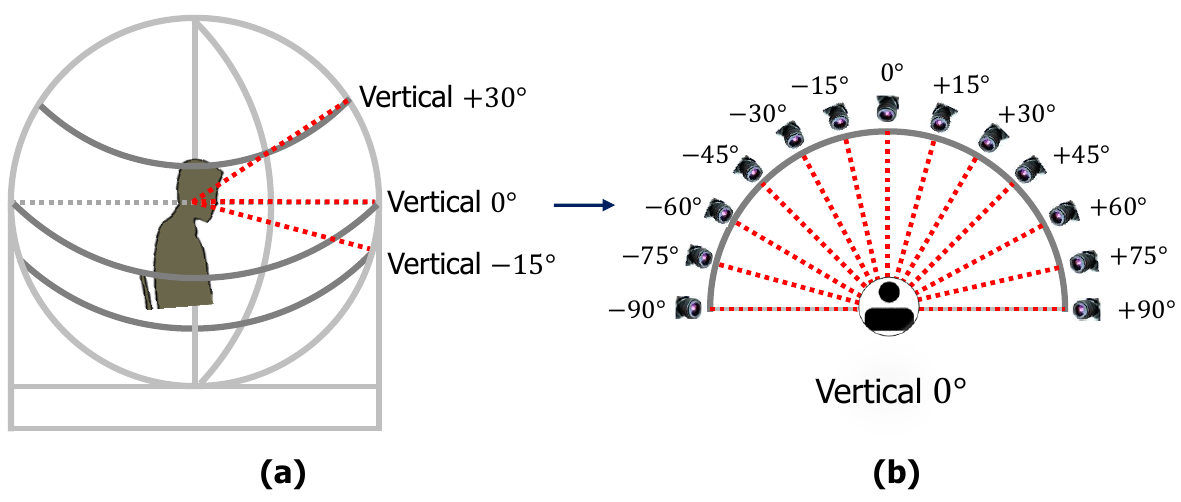}}
  \caption{Positions of the cameras in the capturing device. (a) Vertical axis angle and (b) horizontal axis angle at 0° on the vertical axis.}
  \label{fig:cameras}
\end{figure}

\begin{table*}
  \caption{The configuration of K-FACE database.}
  \label{table:config_kface}
  \includegraphics[width=\linewidth]{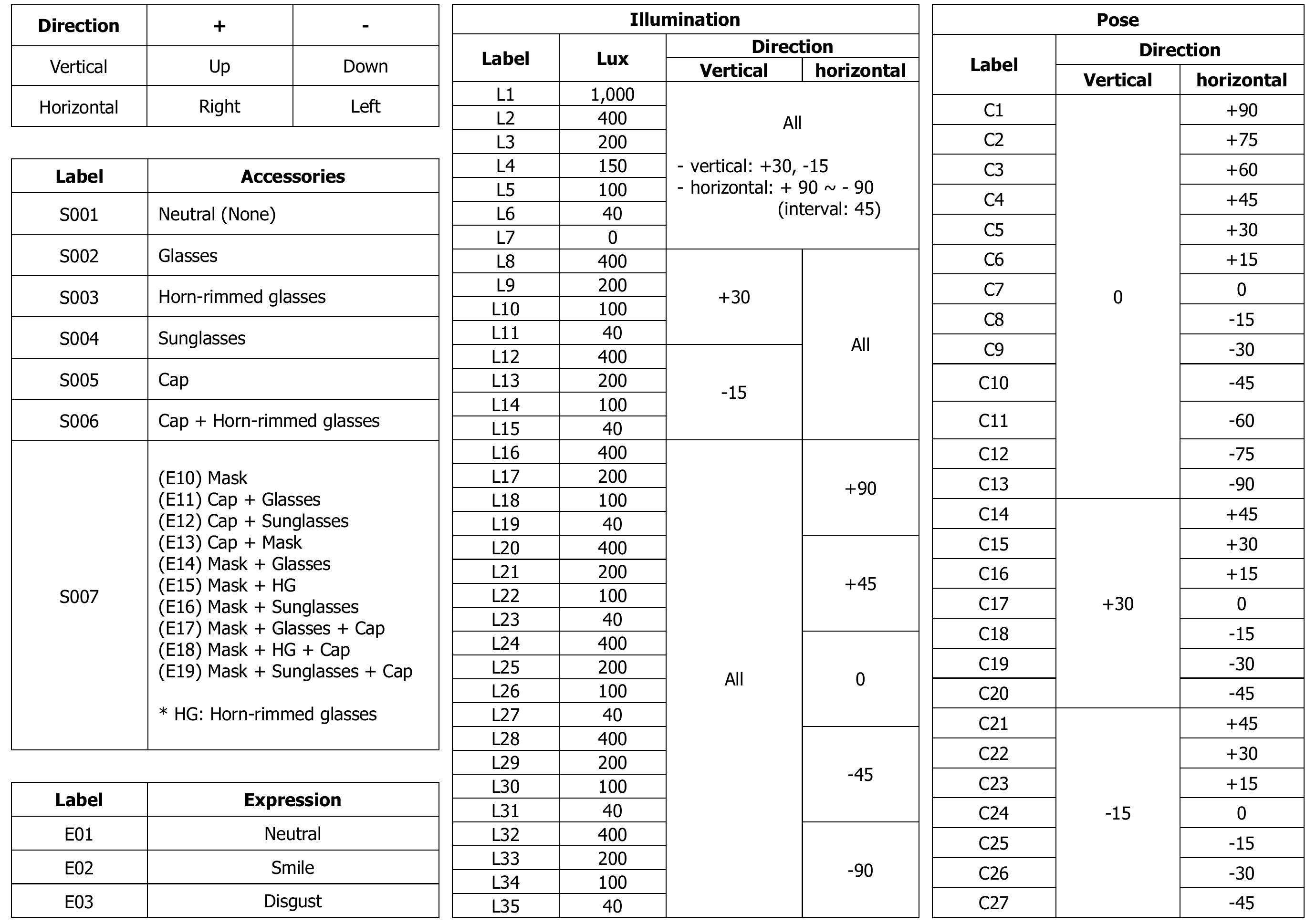}
\end{table*}

\subsection{Capturing Device}
To build the K-FACE database, we designed an elaborate hemispherical system as shown in \figurename{~\ref{fig:device_concept}}. This novel system consists of 27 FHD DSLR cameras and 10 lighting devices installed at regular angular intervals. To acquire images more efficiently and systematically, all cameras were synchronized to acquire data at once. The camera was a Canon EOS 100D, while the resolution of the images was 2592$\times$1728.
As shown in \figurename{~\ref{fig:cameras}}, based on the vertical axis, three hemisphere bars were placed at $0\,^{\circ}$ in front of the face,  $+30\,^{\circ}$, and $-15\,^{\circ}$. There were 13 cameras on the bar at 0° on the vertical axis. Based on the horizontal axis, the front of the face was at  $0\,^{\circ}$ and the cameras were located at $15\,^{\circ}$ intervals from $-90\,^{\circ}$ to $+90\,^{\circ}$ in the left and right directions. The bars at $+30\,^{\circ}$ and $-15\,^{\circ}$ on the vertical axis contained seven cameras at equal intervals from $-45\,^{\circ}$ to $+45\,^{\circ}$. Therefore, a total of 27 cameras were located at (horizontal axis, vertical axis): ($\pm$90, 0), ($\pm$75, 0), ($\pm$60, 0), ($\pm$45, 0), ($\pm$30, 0), ($\pm$15, 0), (0, 0), ($\pm$45, +30), ($\pm$30, +30), ($\pm$15, +30), (0, +30), ($\pm$45, -15), ($\pm$30, -15), ($\pm$15, -15), and (0, -15). 
Moreover, to obtain images in various lighting environments, a total of 10 lighting devices, which can be individually controlled, were placed at (horizontal axis, vertical axis): ($\pm$90, +30), ($\pm$45, +30), (0, +30), ($\pm$90, -15), ($\pm$45, -15) and (0, -15). As shown in Table~\ref{table:config_kface}, we generated a total of 35 different lighting conditions by adjusting the intensities from zero to 1,000 lux and changing the lighting direction to include partial illumination for shadows on the face. All labels regarding poses, lighting intensities, and directions were provided.


\subsection{Data Acquisition and Annotating}
To avoid data bias, a common problem in public databases, caused by the concentration of data on specific age groups or genders that are simple to obtain, we recruited subjects aged 20–50 years. Using the system described in Section 3.1, images of a subject were then acquired while wearing various accessories such as a cap, two types of glasses, sunglasses, and a mask, in various lighting environments for approximately 2 h. We had constructed the K-FACE database for a long time since 2017~\cite{kface}. The total number of images per subject was 17,550, which included diverse poses and three facial expression changes. Finally, we gathered a total of 17,550,000 images from 1,000 subjects. 

Furthermore, to improve the quality and ensure the sophistication of our database, we cleaned and supplemented the data as soon as they were collected. To this end, a full inspection was performed manually. Subsequently, if damaged data existed (e.g., out of focus, closed eyes, not matching the lighting conditions due to synchronization delays), the data were immediately retaken to maintain the data distribution for all subjects and conditions. To expand the usability of the database, we also provided additional information about the facial landmark positions and bounding boxes. 
\begin{figure}[b]
  \centering
  \includegraphics[width=7.5cm]{{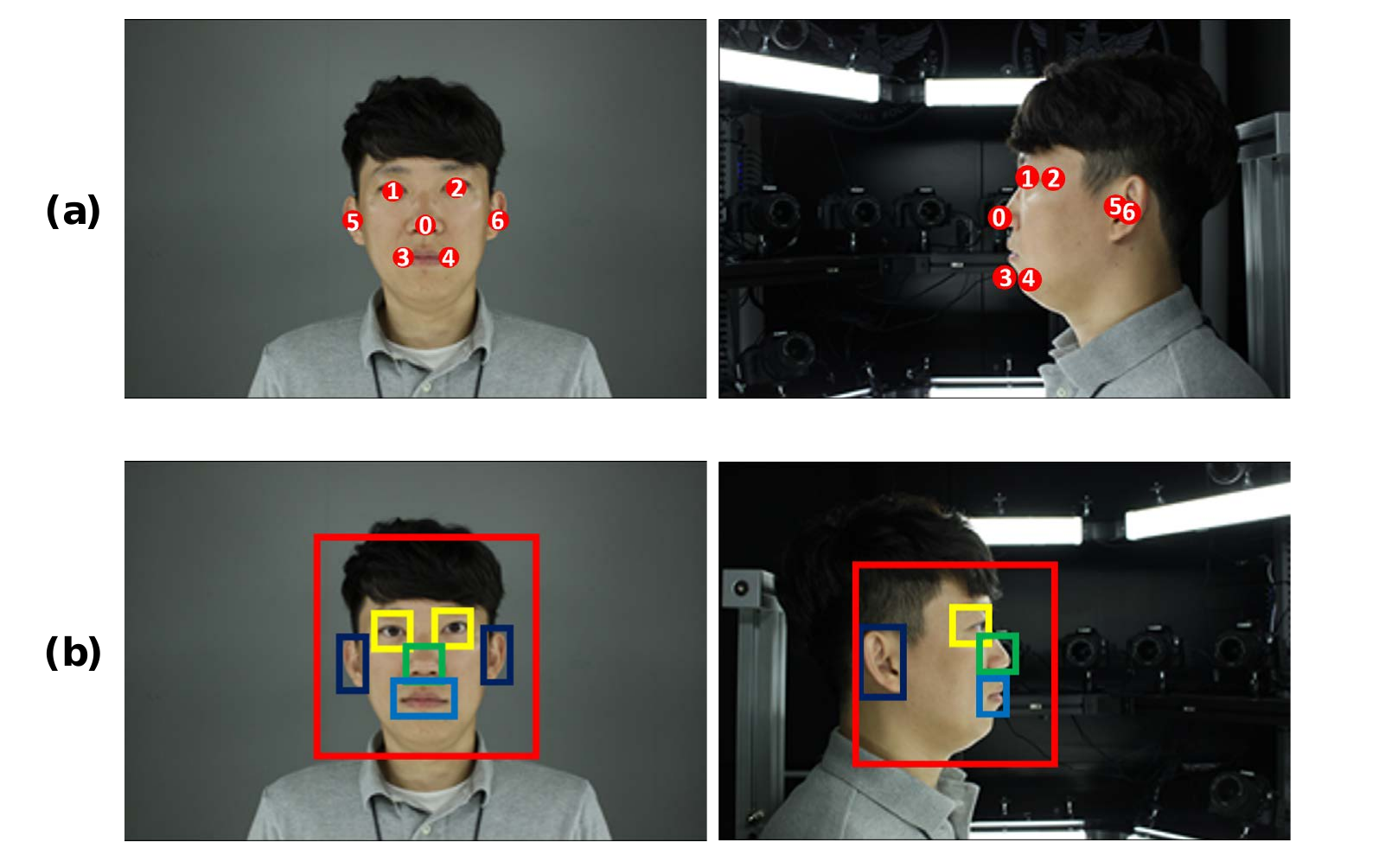}}
  \caption{The positions of (a) facial landmarks and (b) bounding boxes.}
  \label{fig:landmark}
\end{figure}

\begin{figure*}[h]
  \centering
  \includegraphics[width=17cm]{{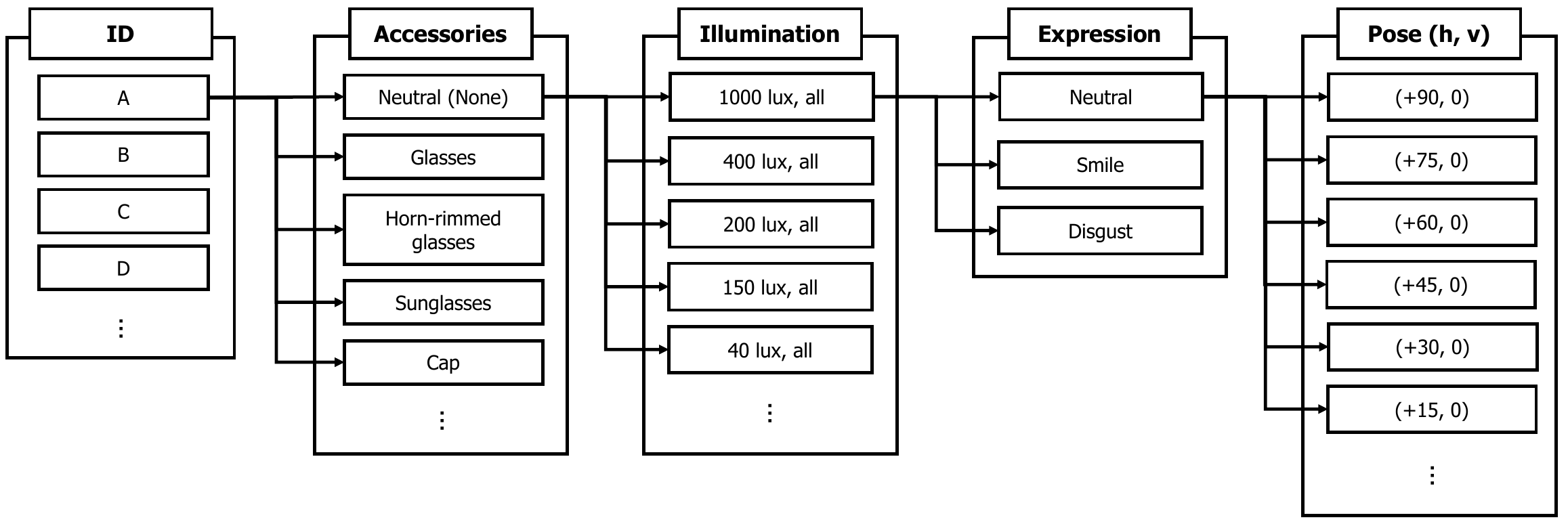}}
  \caption{Structure of the K-FACE database.}
  \label{fig:data_structure}
\end{figure*}

\begin{figure}[h]
  \centering
  \includegraphics[width=8cm]{{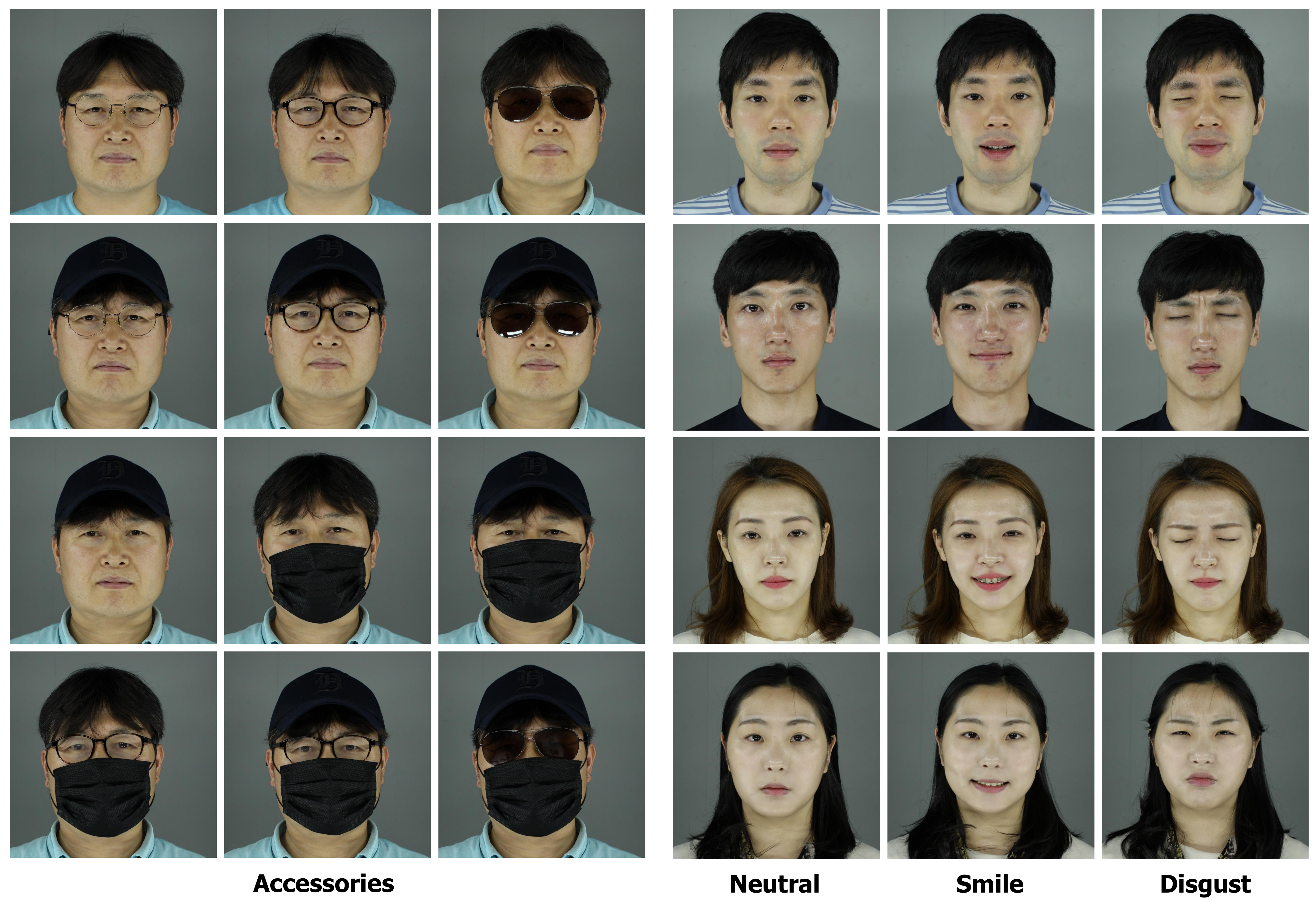}}
  \caption{Sample images of accessories and expressions (neutral, smile, and disgust).}
  \label{fig:acc_expres}
\end{figure}
As shown in \figurename{~\ref{fig:landmark}}, seven coordinates for the facial landmarks are offered: endpoint of the nose, centers of both eyes, left and right angles of the mouth, and centers of both ears. For bounding boxes, seven regions are provided: the entire face, both eyes, both ears, nose, and mouth.

\subsection{Database Statistics}

The K-FACE database was elaborately and systematically constructed while maintaining the ratio of age distribution, gender, and number of images per subject to overcome the limitations in existing public databases due to data imbalance. The data were acquired for those in their 20s to 50s, and the distribution of each age group was the same (approximately 25\%). The proportions of males and females were 49.8\% and 50.2\%, respectively, so that the gender ratio was almost the same. The gender ratios in each age group were also the same. Additionally, all subjects had the same number of images for each condition to enable an accurate statistical analysis of changes in pose, lighting, facial expressions, and accessories.

\begin{figure}[t]
  \centering
  \includegraphics[width=8.5cm]{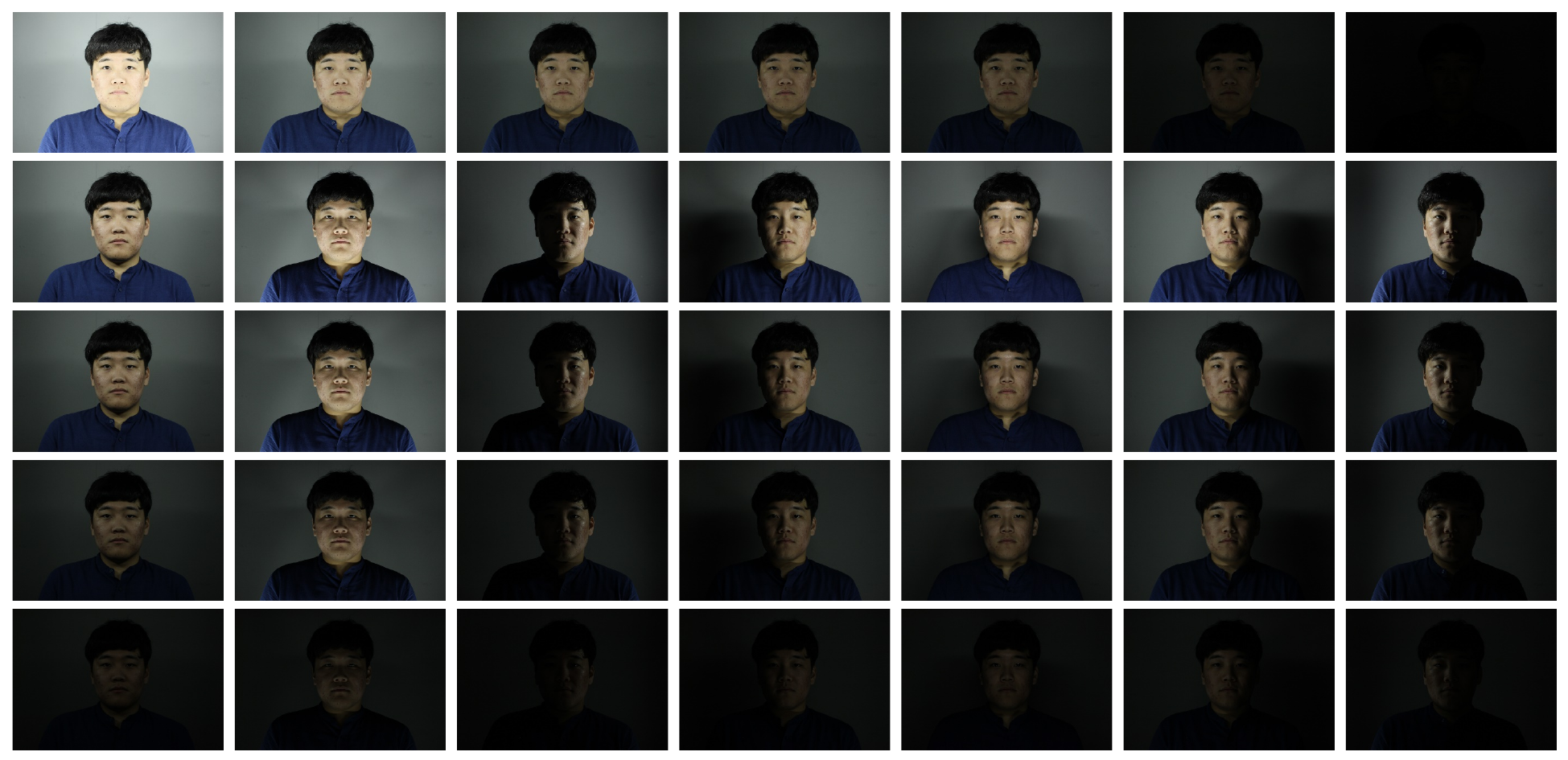}
  \caption{Sample images of lighting changes.}
  \label{fig:illum}
\end{figure}
\begin{figure*}[t]
  \centering
  \includegraphics[width=18cm]{{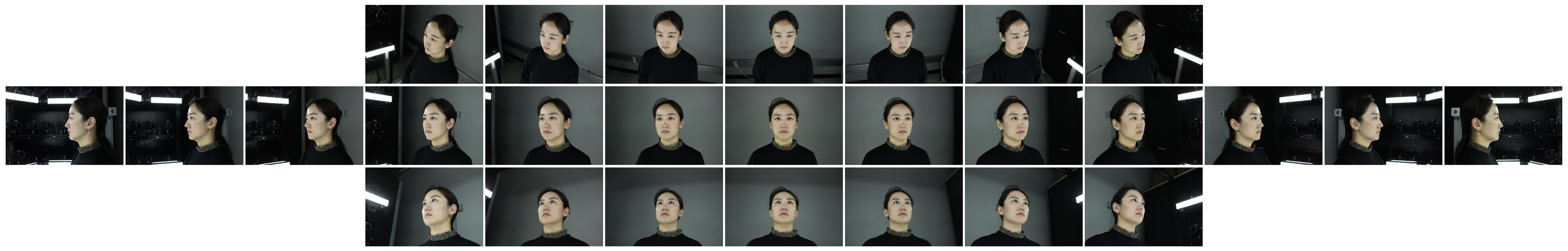}}
  \caption{Sample images of poses.}
  \label{fig:pose}
\end{figure*}

The K-FACE database is structurally well organized to increase usability. As shown in \figurename{~\ref{fig:data_structure}}, it consists of a total of five layers: identities, accessories, illuminations, expressions, and poses. There are a total of seven accessory conditions: neutral (none), where the subject did not wear anything, glasses, horn-rimmed glasses, sunglasses, cap, horn-rimmed glasses and cap, and a combination of some accessories. In the last condition, to respond to the situation where wearing a mask becomes more common due to the spread of the coronavirus disease 2019 (COVID-19), we added a mask and expanded the diversity with a variety of 10 combinations, such as a cap and a mask. The sample images of the accessory conditions are shown in \figurename{~\ref{fig:acc_expres}}; further details are presented in Table~\ref{table:config_kface}. Each accessory condition includes 35 lighting conditions, except for the last accessory condition, which has only two lighting changes. As shown in \figurename{~\ref{fig:illum}}, some extreme illumination cases are included; for instance, when the intensity is close to zero. There are three facial expressions, neutral, smile, and disgust, for each lighting condition except for the last accessory condition. This case was acquired in one expression, neutral. Finally, there were 27 poses per facial expression. The sample images of all expressions and all poses are shown in \figurename{~\ref{fig:acc_expres}} and \figurename{~\ref{fig:pose}}, respectively.

Consequently, the K-FACE database is a uniformly organized database of six accessories, 35 lighting conditions, three facial expressions, and 27 poses. Therefore, the number of images per subject is 17,550: 6 (accessory conditions) × 35 (illumination conditions) × 3 (expressions) × 27 (poses) + 10 (combinations of accessory conditions) × 2 (illumination conditions) × 1 (expression) × 27 (poses). We also provide labels for all information, such as accessory conditions, which are denoted as S001, expressions, illumination changes, and poses, for all images. The notation of each label is shown in \figurename{~\ref{fig:data_structure}}.

\section{Applicable Research Fields}
The proposed K-FACE database can be utilized in all deep-learning research fields related to face images. Among the various research fields, representative challenging research topics that can be addressed by utilizing the K-FACE database are unconstrained face recognition and face age estimation.

\subsection{Unconstrained Face Recognition}
Inspired by the emergence of the large-scale face database and the development of deep-learning-based image classification technology, the accuracy of face recognition has been largely improved to a level that surpasses human recognition performance on some popular benchmarks. Nevertheless, the recognition of face images captured in the real world remains a challenging task, particularly in cases of changing poses, lighting conditions, facial expressions, and wearing accessories. In this respect, the proposed K-FACE database provides not only various face images in a highly balanced distribution, but also accurate labels for all information, such as poses, illumination conditions (intensity and direction), facial expressions, and types of accessories, which is significant in establishing a robust face recognizer in unconstrained environments.

\begin{figure}[t]
  \centering
  \includegraphics[width=8cm]{{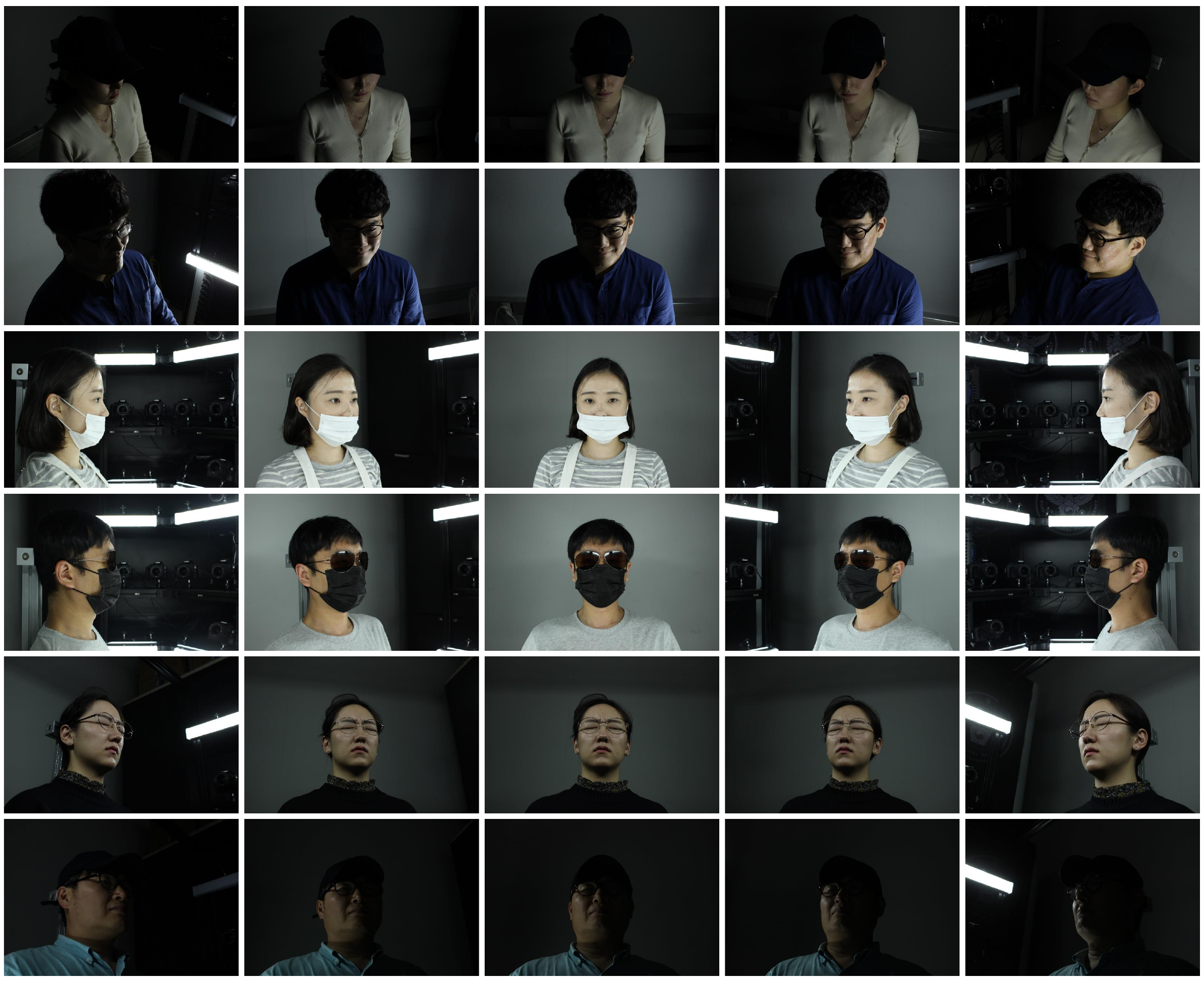}}
  \caption{The sample images for unconstrained face recognition.}
   \vspace{-0.5cm}
  \label{fig:unconst_fr}
\end{figure}

Face image with pose variation is one of the most important factors that can degrade the face recognition performance. To solve this problem, extensive studies have been carried out based on face frontalization, which converts the face image to a frontal view, and pose-invariant face recognition in the real world. As K-FACE includes images in the vertical and horizontal directions, images acquired from the top angle can reproduce public video surveillance such as closed-circuit television and access control systems, while face images acquired from the bottom angle can be used to improve the accuracy in scenarios such as ATM and body cams. 
Regarding the changes in lighting, the face recognition accuracy can be degraded by reflections or shadows caused by an extremely low light intensity or partial illumination. Most existing face databases include only the direction of illumination, while the K-FACE database also includes the intensity of illumination, so that it is possible to study image-based lighting normalization under various conditions and model the lighting environment according to various directions and intensities. Moreover, K-FACE allows researchers to investigate face recognition robust to facial deformation caused by facial expression changes, such as neutral, smile, and disgust, and occlusion caused by wearing a mask, sunglasses, glasses, and a cap. 
In particular, the demand for reliable identification systems for masked faces significantly increases owing to the COVID-19 pandemic~\cite{wang2020masked}. The K-FACE database will enable researchers to develop and evaluate identification algorithms that utilize the upper part of the face or ear recognition. \figurename{~\ref{fig:unconst_fr}} shows sample images in unconstrained environments, including various facial expressions wearing caps, sunglasses, and masks, acquired from different viewing angles under extreme light sources.

\subsection{Face Age Estimation and Aging Simulation}

Age estimation from face images and face aging, also referred to as age synthesis or age progression, has applications in various domains, including age-invariant face recognition, finding missing children, and entertainment~\cite{AgeGAN}. In recent years, numerous face estimation and aging models have been proposed using the face age database and have provided impressive aging results for natural face images~\cite{AgeEst, Agemorph, FFHQ, CACD}. However, the face estimation and aging can still be improved in real cases where many different poses, lighting changes, and occlusions occur. Changes in the above elements largely alter the appearance of the face and distort the morphological facial shape. These facts interfere with the accurate age estimation and often lead to the loss of the original person’s identity in the aging simulations. In this respect, K-FACE enables age estimation applicable in real life as it includes face images with ages of 20 to 60 years acquired at various poses, lightings, facial expressions, and wearing accessories. Through the images provided by K-FACE, it is possible to capture components such as face shape, skin texture, and wrinkles that vary by age group.

\subsection{Others}

In addition to the aforementioned face recognition and age recognition fields, the K-FACE database can be used in various other fields. For instance, to perform an elaborate 3D face restoration using only two-dimensional images without depth information, multiple images acquired from various angles are required. Therefore, in this case, the K-FACE database containing images acquired from various angles is very useful. Furthermore, using information about landmarks and bounding boxes provided in the K-FACE databases, pose estimation, which predicts a pose in the face direction, and face or attribute (such as eyes, nose, and ears) detection can be performed.

\section{Conclusion}

In this paper, we introduced a novel large-scale face database, the K-FACE database, which contains 17,550,000 images of 1,000 subjects. This database was constructed uniformly and systematically by considering the ratio of age groups, the gender ratio, and data distribution per subject. It also includes images acquired in various unconstrained environments, including poses, lighting changes, and accessories. Furthermore, it is possible to perform an accurate statistical analysis of the changes by offering labels for all conditions for all images. Therefore, it is a very valuable database that can be used in various fields, such as unconstrained face recognition, age estimation, and face reconstruction. We also designed a sophisticated device to facilitate the database construction. The quality of the database can be guaranteed through manual data inspection. In future studies, we will carry out various experiments, such as unconstrained face recognition and aging simulation, using state-of-the-art methods to verify the validity and applicability of the K-FACE database.



\section*{Acknowledgement}
This research was funded by R\&D program for Advanced Integrated-intelligence for IDentification (AIID) through the National Research Foundation of Korea (NRF) funded by Ministry of Science and ICT, grant number (2018M3E3A1057288).

{\small
\bibliographystyle{ieee}
\bibliography{egbib}
}

\end{document}